\documentclass{article}

\usepackage[final]{corl_2020} 

\usepackage[utf8]{inputenc} 
\usepackage[T1]{fontenc}    
\usepackage{url}            
\usepackage{booktabs}       
\usepackage{amsfonts}       
\usepackage{nicefrac}       
\usepackage{microtype}      
\usepackage[noend, ruled,vlined]{algorithm2e}
\usepackage{caption}

\usepackage{amsmath}

\usepackage{acronym}
\usepackage[hyperref]{backref}
\usepackage{graphicx}
\usepackage{duckuments}
\usepackage{mleftright}     
\usepackage{enumitem}       
\usepackage{verbatim}       
\usepackage{wrapfig}
\usepackage{subcaption}
\usepackage{floatrow}
\usepackage{mathtools}

\usepackage{booktabs,dcolumn}

\usepackage{tikz}
\usetikzlibrary{decorations.pathreplacing,calc}
\newcommand{\tikzmark}[1]{\tikz[overlay,remember picture] \node (#1) {};}

\newcommand*{\AddNote}[4]{%
    \begin{tikzpicture}[overlay, remember picture]
        \draw [decoration={brace,amplitude=0.5em},decorate, very thick,black]
            ($(#3)!(#1.north)!($(#3)-(0,1)$)$) --  
            ($(#3)!(#2.south)!($(#3)-(0,1)$)$)
                node [align=left, text width=1.5cm, pos=0.5, anchor=west] {#4};
    \end{tikzpicture}
}%

\newcolumntype{d}[1]{D{.}{.}{#1}}

\newcommand*\mcTwo[1]{\multicolumn{2}{l}{#1}}

\acrodef{dlc}[DLC]{Deep Latent Competition}
\acrodef{ilqg}[iLQG]{iterative Linear-Quadratic-Gaussian}
\acrodef{MARL}[MARL]{Multi-Agent Reinforcement Learning}
\acrodef{RL}[RL]{Reinforcement Learning}

\title{Deep Latent Competition: Learning to Race Using Visual Control Policies in Latent Space}

\author{
  Wilko Schwarting$^{*,1}$, Tim Seyde$^{*,1}$, Igor Gilitschenski$^{*,1}$, \\
  {\bf Lucas Liebenwein$^1$, Ryan Sander$^1$, Sertac Karaman$^2$, Daniela Rus$^1$}\\
  $^1$MIT CSAIL, $^2$MIT LIDS, $^*$equal contribution\\
  \texttt{\{wilkos,tseyde,igilitschenski\}@mit.edu}
  
}

\begin{document}
\maketitle

\DeclarePairedDelimiterX{\infdivx}[2]{(}{)}{%
  #1\;\delimsize\|\;#2%
}
\newcommand{\infdiv}{\infdivx}
\DeclarePairedDelimiter{\norm}{\lVert}{\rVert}


\begin{abstract}
Learning competitive behaviors in multi-agent settings such as racing requires long-term reasoning about potential adversarial interactions.
This paper presents Deep Latent Competition (DLC), a novel reinforcement learning algorithm that learns competitive visual control policies through self-play in imagination.
The DLC agent imagines multi-agent interaction sequences in the compact latent space of a learned world model that combines a joint transition function with opponent viewpoint prediction.
Imagined self-play reduces costly sample generation in the real world, while the latent representation enables planning to scale gracefully with observation dimensionality.
%
%
%
%
%
%
We demonstrate the effectiveness of our algorithm in learning competitive behaviors on a novel multi-agent racing benchmark that requires planning from image observations. Code and videos available at \url{https://sites.google.com/view/deep-latent-competition}.
\end{abstract}



\section{Introduction}

\begin{wrapfigure}[26]{r}{0.54\textwidth}
\vspace{-19pt}
\begin{center}
    \includegraphics[width=1.0\textwidth]{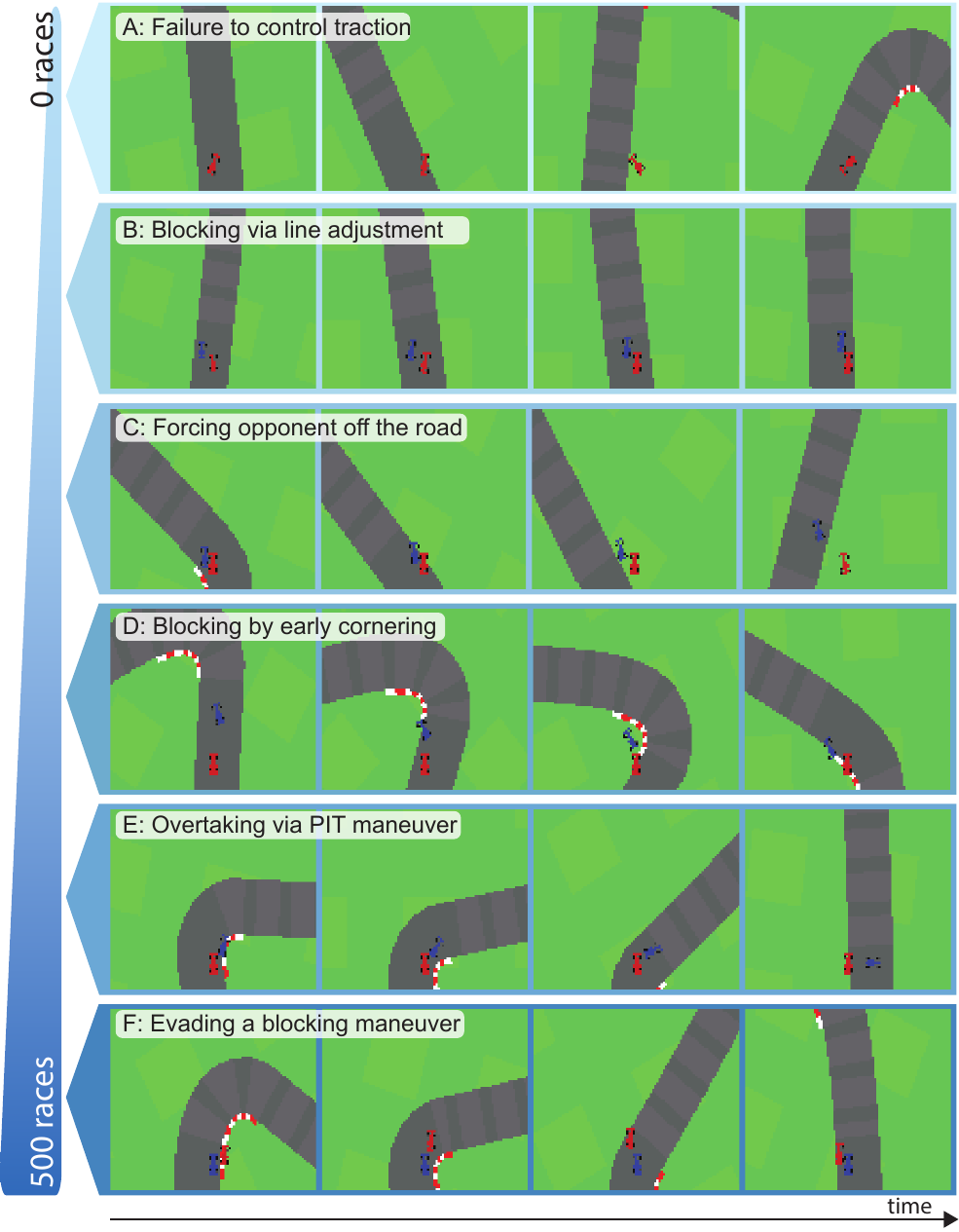}
  \caption{Skill acquisition from experience.}
  \label{fig:skill_acquisition}
  \vspace{0pt}
  \end{center}
  \vspace{-10pt}
\end{wrapfigure}
%
%
Driverless racing is a challenging task promising to push the limits of autonomous navigation as it requires robust operation of the entire driving stack at high speeds. 
Racing becomes particularly challenging when multiple agents simultaneously compete within the same environment. 
This requires both, fast processing of observations and reasoning about opponent behavior. 
By addressing these challenges, racing can provide novel insights for the deployment of autonomous systems.
%
In the context of \ac{MARL}, racing can benchmark competitiveness as it requires reasoning about the interplay between ego actions and opponent behavior. This is particularly challenging when operating under partial observability in high-dimensional input spaces and in the absence of structured priors over environment behavior.



%
%
Recent approaches to autonomous racing tend to assume access to a nominal environment model~\cite{Alcala2020,Carrau2016,Liniger2019}.
Some also leverage game-theoretic frameworks such as iterated best response~\cite{Spica2020,Wang2020,Liniger2020}. 
However, in racing and other real-world multi-agent settings, relevant behavioral and environmental features may be too complex to be captured by MPC or purely game-theoretic approaches that assume perfect knowledge of the state and the system dynamics.
\ac{MARL} provides an alternative framework for modelling rich agent interactions. 
Impressive successes have been demonstrated for large-scale competition in discrete action spaces~\cite{Vinyals2019} with access to privileged ground truth information such as categorized entity lists and  multi-layer maps.
Continuous control applications of \ac{MARL} in competitive settings that only provide image observations are less well-studied.
In particular, combining high-dimensional observations such as images, partial observability of other agents, multi-agent world model learning, and acquiring complex competitive behaviors through self-play is still a challenge.


%
%
We address this research gap by proposing \ac{dlc}, a novel model-based \ac{MARL} method that operates on raw image inputs. 
While this method can be applied to a range of problems, we focus on demonstrating it in the context of two-player racing.
Our approach learns a world model for imagining competitive behavior in latent-space. 
This allows for training agents via imagined self-play such that they can predict opponent behavior and incorporate the expected outcomes of action sequences in their policy selection.
We further learn to predict the belief of other agents purely based on observations from the ego agent's perspective. 
We validate this methodology on a novel multi-agent racing benchmark based on OpenAI Gym~\cite{Brockman2016} that requires the application of continuous visual control policies. 
In summary, this work contains the following contributions:
\begin{itemize}[leftmargin=20pt]
\item A novel model-based reinforcement learning algorithm for \textbf{learning competitive control policies} for multi-agent problems from raw image observations through self-play in latent space
\item A multi-agent world model structure that allows for \textbf{(a) imagination of competing agents' behavior} in a learned latent space and \textbf{(b) estimation of the beliefs of others} from own observations
\item \textbf{Extensive evaluations} in a \textbf{new multi-agent racing benchmark} demonstrating superiority over approaches that do not reason about other agents in imagination
\end{itemize}

\section{Related Work}
\paragraph{Autonomous racing and navigation}
Most recent approaches considering autonomous racing assume knowledge of the underlying dynamics model and use machine learning based techniques for improving said model.
This can be employed in conjunction with a variety of control 
approaches~\cite{Alcala2020,Carrau2016,Liniger2019, schwarting2018planning}.
Game-theoretic methods additionally explicitly model interactions with other agents for decision making~\cite{Schwarting2019,Spica2020,Wang2019,Wang2020,Williams2017,Liniger2020, schwarting2019social}. 
%
These approaches usually do not operate on high-dimensional input spaces and often impose assumptions on the type of interactions.
On the other hand, learning-based end-to-end navigation approaches~\cite{Pomerleau1989, Amini2020, Bansal2019, Bojarski2016, Kendall2018} can operate directly on high-dimensional sensor data but typically involve no inductive biases for considering interactions. 

\paragraph{Multi-agent RL} 

Recently, \ac{MARL} agents have surpassed human-level performance in many multi-agent environments including complex board and card games such as Go~\cite{silver2017mastering}, chess, shogi~\cite{Silver2018}, and Poker~\cite{moravvcik2017deepstack,brown2019superhuman}. \ac{MARL} agents also reached grandmaster-level performance in the real-time strategy game Star Craft II~\cite{Vinyals2019,Vinyals2017}, and showed complex emergent tool use in hide-and-seek~\cite{baker2019emergent}.
We are motivated by the success of recent algorithms for continuous control tasks in cooperative, competitive, and team competition environments~\cite{Li2019,Iqbal2019,Liu2018,bansal2018emergent}.
While these agents leveraged privileged information, such as entity lists, states, and maps, we present an agent that learns competitive strategies directly from raw image observations.
A common thread in \ac{MARL} are self-play auto-curricula~\cite{Lockhart2019,Lanctot2017, Heinrich2015, Heinrich2016}. In contrast, we gain competitiveness from imagined self-play in a learned multi-agent latent world model.

\paragraph{Latent imagination in RL}
Use of neural networks, particularly recurrent neural networks, for modeling the evolution of the environment allowing for \emph{"mental imagination"} has been proposed as early as 1990~\cite{Schmidhuber1990} and recently revisited in~\cite{Ha2018}.
In a similar spirit, variational inference approaches have been combined with the linear-quadratic-regulators for learning to control from raw images~\cite{watter2015embed,zhang2018solar}.
Another line of recent algorithms combines latent (multi-step) imagination with video prediction~\cite{Kaiser2020, Hafner2020Dream, Seyde2020}, achieving state-of-the-art performance on several standard benchmarks. 
We draw inspiration from these ideas and generalize the concept of multi-step latent imagination to multiplayer settings. 

\section{Representing Multi-Agent World Models}
We define competitive visual control in our racing domain as a \ac{MARL} problem. A collection of agents interacts within the environment and learns to optimize their behavior in an effort to maximize individual cumulative reward.
MARL is typically modelled as a Markov game~\cite{shapley1953stochastic, littman1994markov, gmytrasiewicz2005framework}, in which each agent solves a partially observable Markov decision process (POMDP)~\cite{kaelbling1998planning, hauskrecht2000value, sutton2018reinforcement}. 
In the following, we first introduce the general problem definition, outline our approach to imagined self-play and introduce our representation learning approach.

\paragraph{Problem formulation of competitive MARL}
We define the POMDP of agent $i\in \{1,\dots,n\}$ as the tuple $M^i=\langle S, A^i, T, \Omega^i, O^i, R^i \rangle$, where $S$ denotes the set of environment states, $A^i$ the continuous action set, and $T : S \times A^1 \times \cdots \times A^n \to \Pi(S)$ the corresponding transition function with associated probability distribution $\Pi(\cdot)$. 
For each agent $i$, we furthermore define a reward function $R^i: S \times A^i \to \mathbb{R}$, as well as an observation set $\Omega^i$ with corresponding observation function $O^i:S \times A^i \to \Pi(\Omega^i)$. 
In the following, we consider a homogeneous set of agents with identical action and observation spaces as well as reward functions, such that $X^i = X$ for $X = \{A, \Omega, O, R\}$.
We do not assume prior knowledge about the environment.
Thus, the nominal reward function $R$, state transition function $T$, and observation function $O$ are unknown.
Let $q_{\phi}(a^i_t|o^i_{\leq t}, a^i_{< t})$ denote the policy of agent $i$, conditioned only on its own observation-action history, and define the associated expected return over a race of duration $T$ to be $\mathbb{E} \sum_{t=1}^{T} r^i_t$. The objective is then to develop an agent that maximizes the expected return in the absence of prior knowledge about nominal environment and opponent behavior.
This defines an extensive-form game as the agents are competing for reward over multiple timesteps, while only receiving instantaneous environment feedback via high-dimensional observations $o_t^i$ and scalar rewards $r_t^i$.

\begin{figure}
     \centering
     \begin{subfigure}[b]{0.325\textwidth}
         \centering
         \includegraphics[width=\textwidth]{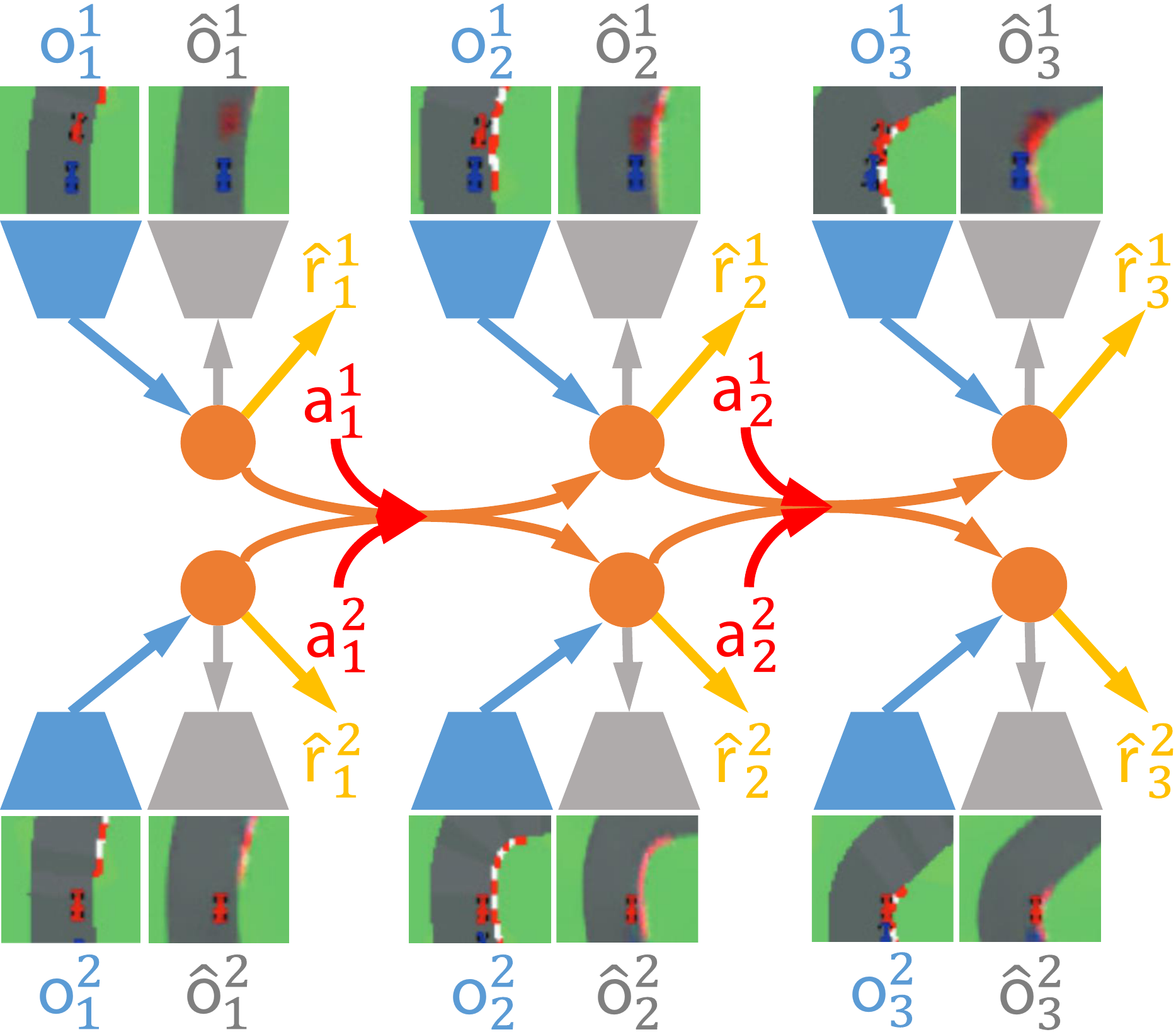}
         \caption{Model learning}
         \label{fig:schematic_dynamics}
     \end{subfigure}
     \hfill
     \begin{subfigure}[b]{0.325\textwidth}
         \centering
         \includegraphics[width=\textwidth]{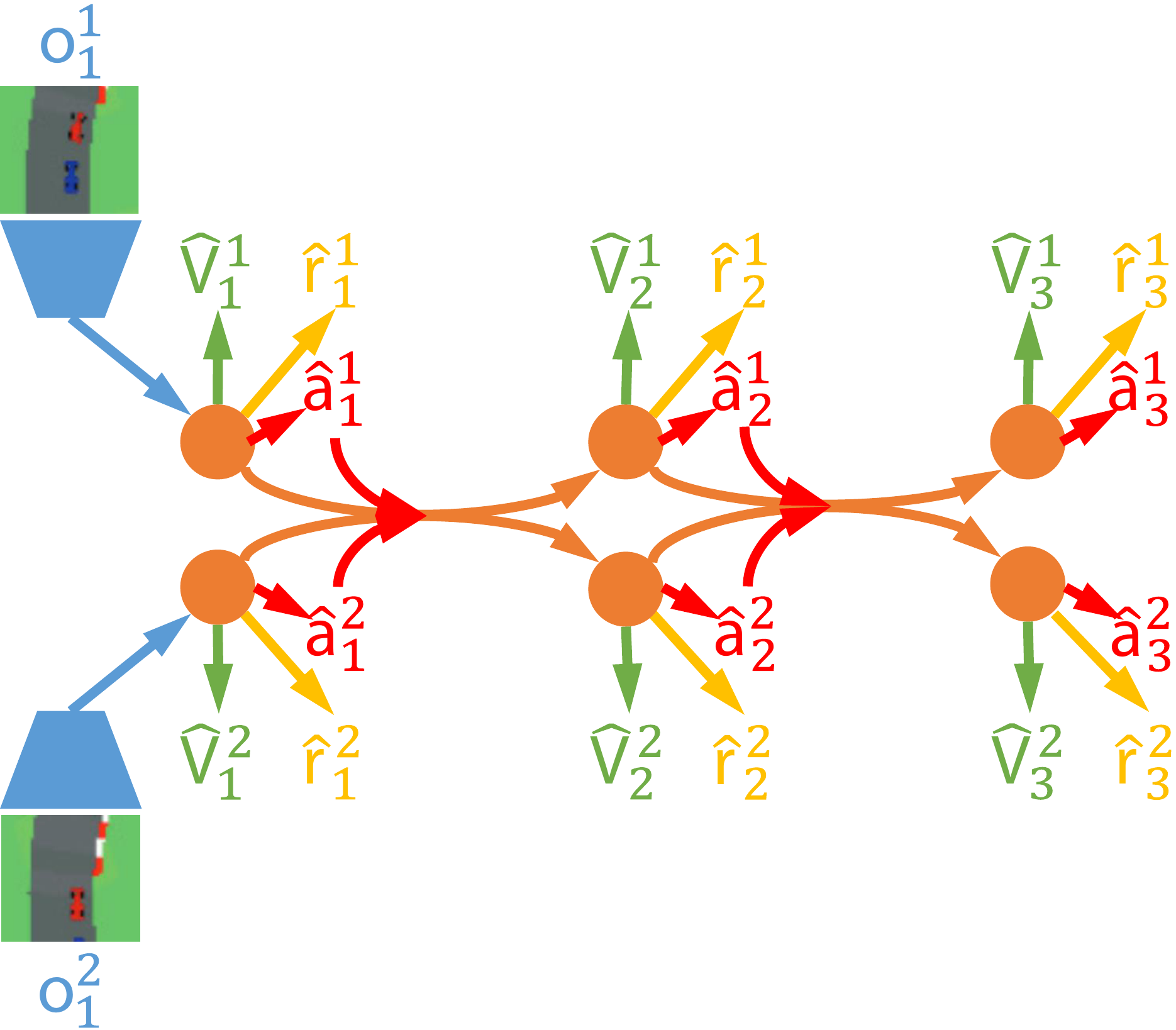}
         \caption{Behavior optimization}
         \label{fig:schematic_imagination}
     \end{subfigure}
     \hfill
     \begin{subfigure}[b]{0.26\textwidth}
         \centering
         \includegraphics[width=\textwidth]{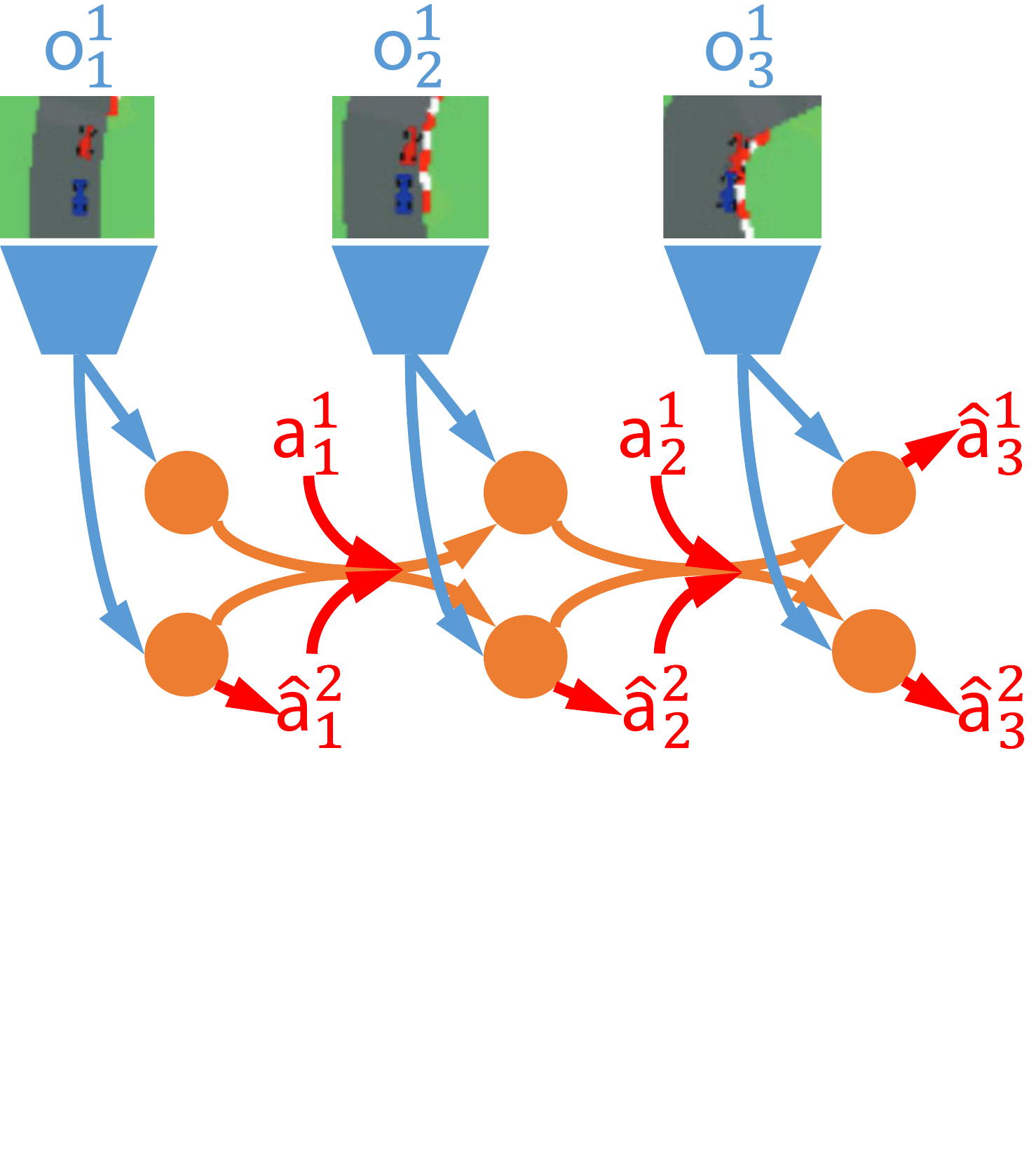}
         \caption{Environment interaction}
         \label{fig:schematic_act}
     \end{subfigure}
        \caption{(a) The agent learns to encode observations into separate latent states for each agent based on reconstruction and reward prediction. The learned transition model propagates all agents' latent states jointly. (b) Self-play in imagination: The agent predicts state values and optimizes actions that maximize future returns by propagating gradients back through imagined game trajectories. The agent's competitiveness continuously improves through self-play. (c) The agent estimates the own and other agents' current state based on an encoding of own historic observations only and predicts the actions of the other agent and the own actions to be executed in the environment. True observations and actions of other agents are not available.}
        \label{fig:schematic_overall}
\end{figure}

\paragraph{Learning through imagined self-play}
Model-based reinforcement learning consists of the tasks of (a) model learning, (b) behavior optimization, and (c) environment interaction.
Model-based \ac{MARL} extends model learning to include predictions of other agents' behavior, while behavior optimization needs to account for competitive fitness.
The training proceeds centralized. While the learning algorithm has access to the observation-action histories of all agents, the deployment is decentralized, providing each agent only with their individual observation-action history.
As detailed in Figure~\ref{fig:schematic_overall}, our algorithm iteratively executes the following:
\begin{itemize}[leftmargin=20pt]
    \item Learning a world model consisting of the joint dynamics and reward function based on previous experience of all agents, see~Figure~\ref{fig:schematic_dynamics}. 
    Learning to predict how the world evolves conditioned on own and expected opponent actions enables each agent to imagine the outcome of games without requiring additional real-world experience.
    %
    \item Learning action and value function models through policy iteration on imagined model rollouts. The agents interact with their adversaries through imagined self-play and acquire increasingly competitive behaviors without the necessity for execution in the real world, see~Figure~\ref{fig:schematic_imagination}.
    \item Competing in the real world to collect novel experience and judge the performance of the current behavior. Each agent only has access to their own observation-action history and performance indirectly depends on how well the states of opponents are being estimated, see~Figure~\ref{fig:schematic_act}.
\end{itemize}

\paragraph{Representing multi-agent world models}
Leveraging a multi-agent world model accelerates learning through imagined self-play. The agent optimizes its behavior by simulating interactions with its opponents in the environment without execution in the real world.
In contrast to single-agent world models \cite{watter2015embed, zhang2018solar, hafner2018learning}, this requires explicit representation of the state and action of each agent, as well as a mechanism for an agent to predict behavior of another agent.
%
In the following, we will consider a two-player game ($n=2$) and extend the representation model formulation provided in \cite{Hafner2020Dream} to yield
\begin{equation}
    \begin{aligned}
        & \text{Representation model:} && \qquad  p_\theta \left(s_{t}^1, s_{t}^2 | s_{t-1}^1, s_{t-1}^2, a_{t-1}^1, a_{t-1}^2, o_t^1, o_t^2\right) \\
        & \text{Transition model:} && \qquad  q_\theta \left(s_{t}^1, s_{t}^2 | s_{t-1}^1, s_{t-1}^2, a_{t-1}^1, a_{t-1}^2\right) \\
        & \text{Encoder model:} && \qquad  q_\theta \left(z_{t}^i, \tilde{z}_{t}^{\neg i}\right | o_{t}^i)\\
        & \text{Observation model:} && \qquad  q_\theta \left(o_{t}^i | s_{t}^i\right)\\
        & \text{Reward model:} && \qquad q_\theta \left(r_{t}^i | s_{t}^i\right) \\
    \end{aligned}
    \label{eq:models_representation}
\end{equation}
where $p$ and $q$ denote distributions in latent space, with $\theta$ as their joint parameterization.
The representation model encodes observations $(o_t^1, o_t^2)$ into Markovian model states $(s_t^1, s_t^2)$ \cite{Hafner2020Dream}, which are propagated under a joint transition function to predict future model states $(s_{t+1}^1, s_{t+1}^2)$.
We explicitly provide the underlying encoder model to emphasize that each agent not only learns an embedding corresponding to its own viewpoint, $z_{t}^i$, but additionally learns to predict embeddings of the opponent, $\tilde{z}_{t}^{\neg i}$. 
This is crucial during deployment, as ground truth observations and actions of opponents are not available.
Instead, we leverage our predicted embeddings $\tilde{z}_{t}^{\neg i}$ in conjunction with an action model $q_\phi\left(a_{t}^{\neg i} | s_{t}^{\neg i}\right)$ and employ a slightly modified representation model $p_\theta \left(s_{t}^i, s_{t}^{\neg i} | s_{t-1}^i, s_{t-1}^{\neg i}, a_{t-1}^i, a_{t-1}^{\neg i}, o_t^i\right)$ that is only conditioned on the observation of the ego agent, $o_t^1$.
The encoder model is part of the representation model and embeds observations into embedding states based on which model states are then generated.
For each agent, we furthermore define an individual observation model $q_\theta\left(o_{t}^i | s_{t}^i\right)$ and reward model $q_\theta\left(r_{t}^i | s_{t}^i\right)$.
The underlying architectures follow \citep{hafner2018learning, Hafner2020Dream}, where the transition model is represented by a recurrent state space model (RSSM), the encoder and observation models by a convolutional neural network (CNN) and transposed CNN, respectively, and the reward model by a dense neural network.
Training of these models then proceeds centralized, such that the learning algorithm is given access to the interaction histories of each agent.

\section{Learning to Compete by Imagined Self-Play}
%
%
Our proposed algorithm learns a world model from ground truth data, based on which behavior is refined through imagined self-play. In the following, we introduce the objectives for the two stages.

\paragraph{Representation learning}
The representation learning objective combines image reconstruction with reward prediction in order to discover latent spaces that not only offer compact representations of environment states but further facilitate prediction of associated trajectory performance.
Because agents are actively competing for reward, their states are propagated jointly through the transition model with each agent learning to predict relevant opponent states.
The observation model then not only provides reconstruction signals for the ego perspective via the true ego state $s_t^i$, but also for the opponent perspective via the predicted opponent state $\tilde{s}_t^{\neg i}$.
Following \citep{Hafner2020Dream}, the models introduced in Eq.~(\ref{eq:models_representation}) are optimized to maximize a reformulation of their variational lower bound objective:
\begin{equation}
    \begin{aligned}
        \textstyle J_{M, \hat{s}} &= \mathbb{E}_{\mathcal{D}} \mleft(\textstyle\sum_{t} \mleft(J_{O, t} + J_{R, t} + J_{D, t}\mright)\mright) \\
         \textstyle J_{O, t} &= \ln{q(o_t^1 | \hat{s}_t^1)} + \ln{q(o_t^2 | \hat{s}_t^2)}\\
         \textstyle J_{R, t} &= \ln{q(r_t^1 | \hat{s}_t^1)} + \ln{q(r_t^2 | \hat{s}_t^2)} \\
         \textstyle J_{D, t} &= -\beta \, \text{KL} \infdiv{p(\hat{s}_t^1, \hat{s}_t^2 | \hat{s}_{t-1}^1, \hat{s}_{t-1}^2,a_{t-1}^1, a_{t-1}^2, o_{t}^1, o_{t}^2)}{q (\hat{s}_{t}^1, \hat{s}_{t}^2 | \hat{s}_{t-1}^1, \hat{s}_{t-1}^2, a_{t-1}^1, a_{t-1}^2)},
    \end{aligned}
    \label{eq:objective_representation}
\end{equation}
where the general model state $\hat{s}$ may either originate from ground truth embeddings $s$ or their predictions $\tilde{s}$. In practice, we optimize a linear combination of $J_{M, \hat{s}}$ with $\hat{s} = \{(s_t^1, s_t^2), (s_t^1, \tilde{s}_t^2), (\tilde{s}_t^1, s_t^2)\}$.
This enables learning of latent representations conducive to solving the task via the ground truth embeddings, while constraining predicted embeddings to sensible representations within that space. 

\paragraph{Behavior learning}
The behavior learning objective optimizes for competitive fitness by maximizing the expected return of the action model $q_{\phi}\mleft(a_{t}^i | s_{t}^i\mright)$ over a $T$-step race.
The agent can imagine outcomes of potential interaction sequences by leveraging the learned transition model, therefore bypassing execution in the real world through imagined self-play. 
Generating entire race sequences can be computationally prohibitive and we follow \cite{Hafner2020Dream} in complementing finite horizon model rollouts with predictions from a value model $v_{\psi}\mleft(s_{t}^i\mright)$ in order to approximate returns corresponding to an extensive form race.
The action and value model are then trained jointly using policy iteration on the objectives
\begin{equation}
    \begin{aligned}
        \textstyle \max_{\phi} \mathbb{E}_{q_{\theta}, q_{\phi}}\mleft(\sum_{\tau=t}^{t+H} V_{\lambda}\mleft(s_{\tau}\mright)\mright), \qquad
        \min_{\psi} \mathbb{E}_{q_{\theta}, q_{\phi}}\mleft(\sum_{\tau=t}^{t+H} \lVert v_{\psi}\mleft(s_{\tau}\mright) - V_{\lambda} \mleft(s_{\tau}\mright)\rVert ^{2}\mright),
    \end{aligned}
    \label{eq:dreamer_objectives}
\end{equation}
where $V_{\lambda}\mleft(s_{\tau}\mright)$ represents an exponentially recency-weighted average of the $k$-step value estimates $V_{N}^{k}\mleft(s_{\tau}\mright)$ to stabilize the learning \cite{sutton2018reinforcement}. We provide the corresponding value function definitions as
\begin{equation}
    \begin{aligned}
         V_{\lambda}\mleft(s_{\tau}\mright) &= \mleft(1 - \lambda\mright)     \textstyle \sum_{n=1}^{H-1} \lambda^{n-1} V_{N}^{n}\mleft(s_{\tau}\mright) + \lambda^{H-1} V_{N}^{H} \mleft(s_{\tau}\mright), \\
         V_{N}^{k}\mleft(s_{\tau}\mright) &= \mathbb{E}_{q_{\theta}, q_{\phi}} \mleft(\textstyle \sum_{n=\tau}^{h-1} \gamma^{n-\tau} r_{n} + \gamma^{h-\tau} v_{\psi} \mleft(s_{h}\mright) \mright),
    \end{aligned}
    \label{eq:value_estimates}
\end{equation}
where $h = \min \mleft(\tau+k, t+H\mright)$. This process leverages imagined trajectories $\{\mleft(s_{\tau}^1, s_{\tau}^2, a_{\tau}^1, a_{\tau}^2\mright)\}_{\tau=t}^{t+H}$ over a horizon of $H$ starting from each timestep $t$ within the sampled batch sequence, where opponent behavior is estimated by querying the action model with predicted latent viewpoints as $q_{\phi}\mleft(a_{t}^{\neg i} | \tilde{s}_{t}^{\neg i}\mright)$.
The resulting algorithm optimizes policies by back-propagating analytic value gradients of imagined self-play trajectories through the learned multi-agent world model.

\paragraph{Deep latent competition}
The resulting algorithm is provided as pseudocode in Algorithm~\ref{alg:dlc}.
It runs for $K$ episodes and proceeds in two phases: during the online phase, data is collected from a $T$-step race with each agent only having access to their respective observations and relying on predicted opponent behavior. 
During the offline phase, representation learning and policy iteration propagate information into the world model and the action model based on the interaction histories of all agents. 
To this end, batch sequences of length $L$ are sampled from replay memory $\mathcal{D}$ to serve as targets for representation learning. 
The behavior is then refined in simulation based on rollout trajectories of length $H$ starting from the ground truth samples which are used in generating the value estimates according to Eq.~\ref{eq:value_estimates}.
Here, we set the underlying parameters to $T=1000$, $L=50$ and $H=15$.
All models are then optimized on the previous objectives with the Adam optimizer \citep{kingma2014adam}.

\begin{algorithm}[ht!]
    \SetAlgoLined\DontPrintSemicolon
    \SetKwInOut{Initialize}{Initialize}
    \SetKwInput{Set}{Set}
    \SetKwFunction{dlc}{dlc}
    \SetKwFunction{rollout}{rollout}
    \SetKwProg{myalg}{Algorithm}{}{}
        \nl \Initialize{ model parameters $\{\theta, \psi, \phi\}$ randomly; memory $\mathcal{D}$ with $5$ random episodes}
        \nl \For{episode $k \leftarrow 1$ \KwTo $K$}{
            \nl \For{timestep $t \leftarrow 1$ \KwTo $T$ \tikzmark{top1}}{
                \nl Observe $o_t^1$ and predict embeddings $z_{t}^1, \tilde{z}_{t}^{2} \sim q_{\theta}\mleft(z_{t}^1, \tilde{z}_{t}^{2} | o_t^1 \mright)$\;
                \nl Propagate states $s_{t}^1, \tilde{s}_{t}^{2} \sim p_{\theta}\mleft(s_t^1, \tilde{s}_t^{2} | s_{t-1}^1, \tilde{s}_{t-1}^{2}, a_{t-1}^1, \tilde{a}_{t-1}^{2}, z_{t}^1, \tilde{z}_{t}^{2}\mright)$\; 
                \nl Generate action $a_t^1 \sim q_{\phi}\mleft(a_t^1 | s_t^1\mright)$, predicted response $\tilde{a}_{t}^{2} \sim q_{\phi}\mleft(\tilde{a}_{t}^{2}  | \tilde{s}_{t}^{2} \mright)$ \qquad \qquad \quad \tikzmark{right1}\;
                \nl Execute $a_t^1$ in the environment\;
            }
            \nl Add episode transitions $\{\mleft(o_t^1, o_t^2, a_t^1, a_t^2, r_t^1, r_t^2\mright)\}_{t=1}^{T}$ to memory $\mathcal{D}$ \tikzmark{bottom1}\;
            \nl \For{trainstep $s \leftarrow 1$ \KwTo $S$ \tikzmark{top2}}{
                \nl Sample batch of sequences $\{\mleft(o_t^1, o_t^2, a_t^1, a_t^2, r_t^1, r_t^2\mright)\}_{t=b}^{b+L} \sim \mathcal{D}$ \tikzmark{top3}\;
                \nl Use the encoder model to predict embeddings $z_{t}^1, \tilde{z}_{t}^1, z_{t}^{2}, \tilde{z}_{t}^{2}$\;
                \nl Use the representation model to predict states $s_{t}^1, \tilde{s}_{t}^1, s_{t}^{2}, \tilde{s}_{t}^{2}$ \;
                \nl Update $\theta$ via representation learning on $\{\mleft(s_{t}^1, s_{t}^2\mright), \mleft(s_{t}^1, \tilde{s}_{t}^2\mright), \mleft(\tilde{s}_{t}^1, s_{t}^2\mright)\}$ \tikzmark{right2} \tikzmark{bottom3}\;
                \nl Compute value estimates $V_{\lambda}^1\mleft(s_{\tau}^1\mright), V_{\lambda}^2\mleft(s_{\tau}^2\mright) \leftarrow$ \rollout{$s_t^1, s_t^2, H$} \tikzmark{top4}\;
                \nl Update $\phi$ and $\psi$ based on Eq.~(\ref{eq:dreamer_objectives}) for all targets $\{V_{\lambda}^1\mleft(s_{\tau}^1\mright), V_{\lambda}^2\mleft(s_{\tau}^2\mright)\}$ \tikzmark{bottom2} \tikzmark{bottom4}\;
            }
        }
    \AddNote{top1}{bottom1}{right1}{\hspace{0.3em} Online}
    \AddNote{top2}{bottom2}{right1}{\hspace{0.3em} Offline}
    \AddNote{top3}{bottom3}{right2}{\hspace{0.3em} Model \\ \hspace{0.3em} Update}
    \AddNote{top4}{bottom4}{right2}{\hspace{0.3em} Policy \\ \hspace{0.3em} Update}
\caption{Deep Latent Competition (DLC)}%
\label{alg:dlc}%
\end{algorithm}%
%




\section{Latent Racing Experiments}
We demonstrate the ability of DLC to learn competitive visual control policies in a novel multi-agent racing environment, and compare performance against baselines to highlight the importance of both the joint transition model and the learned observer. We further highlight the learned representation model's capability of predicting opponent viewpoints from ego observations.
\begin{figure}
\begin{center}
    \includegraphics[width=1.0\linewidth]{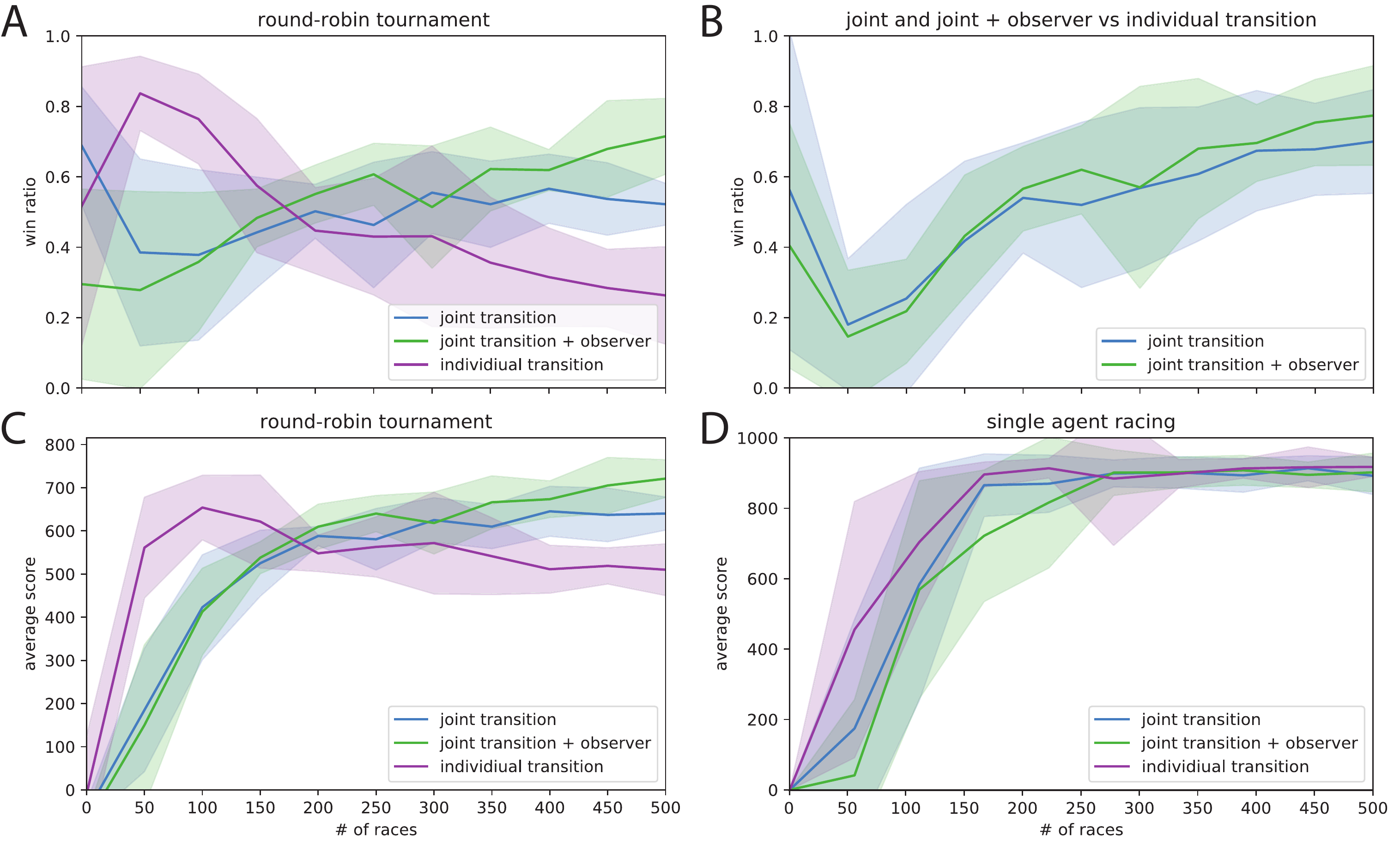}
\end{center}
\caption{\textbf{A}: Win ratio of all agents in a round-robin tournament. \textbf{B}: Agents with joint and joint + observer transition compete directly against an agent with individual transition function.
\textbf{C}: Average score of all agents in a round-robin tournament. \textbf{D}: Single agent racing performance disentangles general skill learning from learning to compete through interaction.}
\label{fig:competition_combined}
\end{figure}

\paragraph{Racing environment}
%
We propose \texttt{MultiCarRacing-v0}, a novel multi-agent racing environment for learning competitive visual control policies\footnote{Code available at \url{https://github.com/igilitschenski/multi_car_racing}}.
The environment extends the Gym task \texttt{CarRacing-v0}~\cite{Brockman2016} and provides each agent with top-down 96x96 pixel image observations from their ego perspective based on which continuous control inputs need to be selected.
The viewpoint is motivated by recent results in the context of driving~\cite{Drews2017,Chen2020, Mani2020} and holds the promise of future deployment on physical platforms. 
%
%
The environment allows for differentiating between skillful and competitive driving. While the former is the basis for high-performance racing, learning to beat a skillful opponent is a far greater challenge. 
Interactions between agents are sparse but information-rich: only when agents collide, push, or block each other do they directly impact each others state.
%
 %

\paragraph{Dynamics and rewards}
The vehicles in the environment exhibit slip and collision dynamics. Breaking or accelerating too hard induces skidding and in combination with steering causes substantial understeering or oversteering (drifting). Similarly, moving off the track lowers available friction.
Collision dynamics allow for elaborate interaction strategies during the race: pushing other agents off the track, blocking overtaking attempts, or turning an opponent sideways via a PIT maneuver. We show examples in Figure~\ref{fig:skill_acquisition}.
%
The reward mapping follows~\texttt{CarRacing-v0}, in that each agent incurs a loss of $-0.1$ per timestep and receives a reward for each visited tile along the track.
%
We incentivize competition by discounting rewards based on visitation order: the first agent to visit a track tile is rewarded with $+1000/N$, while the second agent receives $+500/N$ (on an $N$-tile track).
\begin{figure}[ht!]
\begin{center}
    \includegraphics[width=1.0\linewidth]{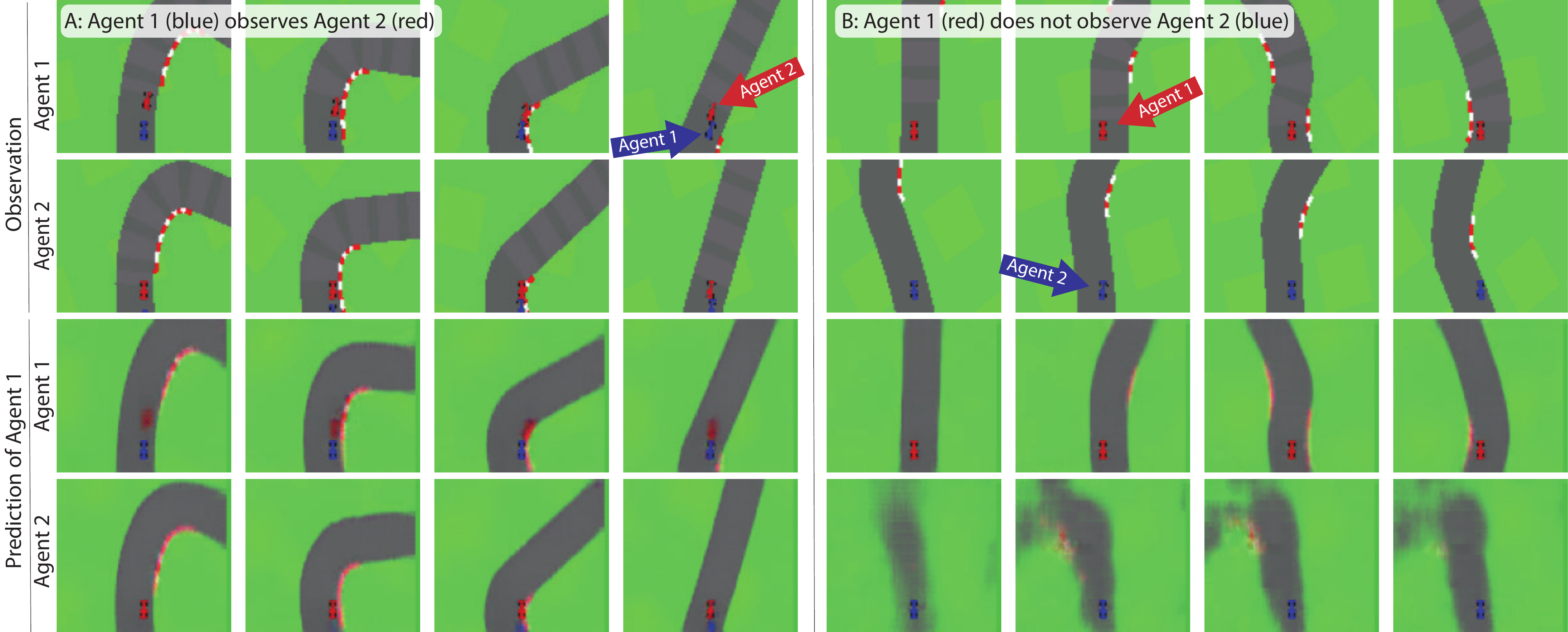}
\end{center}
\caption{\textbf{Closed loop prediction: }\textbf{A}: Agent 1 observes the opponent and is therefore able to reconstruct both views. \textbf{B}: 1 is unable to observe 2 and reconstructs their view with high uncertainty.}
\label{fig:observer}
\end{figure}

\paragraph{Benchmarking performance}
%
The algorithm presented in this work, DLC, learns a multi-agent world model that enables imagined self-play by combining the underlying transition function with an observer capable of predicting opponent latent states. We refer to this method as \textit{joint transition + observer} and compare performance against two baselines.
%
%
The first is \textit{joint transition}, which propagates ground truth latent states of both agents and allows for assessing performance of the observer.
%
%
The second is \textit{individual transition}, which propagates ground truth latent states individually and highlights the added value of a joint transition model.
Each method is trained on 500 races. We evaluate the resulting performance in a round-robin tournament of 100 races for each pairing (300 races per tournament) at multiple stages of training progress. We repeat this for $5$ random seeds.

Figure~\ref{fig:competition_combined}A shows the resulting win-ratio and Figure~\ref{fig:competition_combined}B the average score for the round-robin tournament.
The  representation learning problem for \textit{individual transition} is easier as the learned transition function does not need to disentangle how the states and actions of both agents affect the transition. Therefore, the \textit{individual transition} baseline starts off strong. 
However, after 200 races both joint transition methods have learned to compete more effectively through imagined self-play. 
%
%
The \textit{individual transition} baseline can not leverage this effect, as agent actions do not affect opponents states during imagined rollouts and self-play may only occur in the real world.
As visible in Figure~\ref{fig:competition_combined}C, after 500 races both joint transition approaches win 70-80\% of races against the \textit{individual transition} baseline. 
Furthermore, both joint transition methods perform similarly, suggesting that the learned observer is capable of predicting opponent latent states in a way that is sufficient in order to induce learning of competitive behaviors.

The single agent racing performance, see~Figure~\ref{fig:competition_combined}D, further allows us to disentangle general skill learning from learning to compete. It confirms that the \textit{individual transition} baseline is able to acquire general racing skills faster. Similarly, the joint transition methods do not outperform the \textit{individual transition} baseline in the single agent racing case, such that their performance increase for the multi-agent setting in Figure~\ref{fig:competition_combined}A can be explained by their increased ability to compete.
%

\paragraph{Predicting the opponents latent state from ego observations}
Given an agent's own history of observations and controls, we can estimate the compact latent state and it's associated reconstruction to visualize the representation model's understanding of the world. More interestingly, we can create a reconstruction of the predicted opponent's latent state to visualize how well the agent can infer the opponent view. Given a predicted opponent state we can predict their actions based on the learned policy. Figure~\ref{fig:observer} provides two scenarios with reconstructions based only on observations of agent 1. 

In Figure~\ref{fig:observer}A, the opponent is within the field of view of agent 1. The reconstructions are focused and close to the ground truth for both the ego and opponent viewpoints, indicating that the latent state is accurate and well understood.
In Figure~\ref{fig:observer}B, agent 1 never observes its opponent and is unable to accurately predict their view. However, instead of predicting only noise, agent 2 is imagined to drive on a straight road with the possibility of an upcoming turn. This is a sensible prediction given that agent 2 is currently not observable and in turn will not immediately affect the motion plan of agent 1.

\begin{figure}[t!]
\begin{center}
    \includegraphics[width=1.0\linewidth]{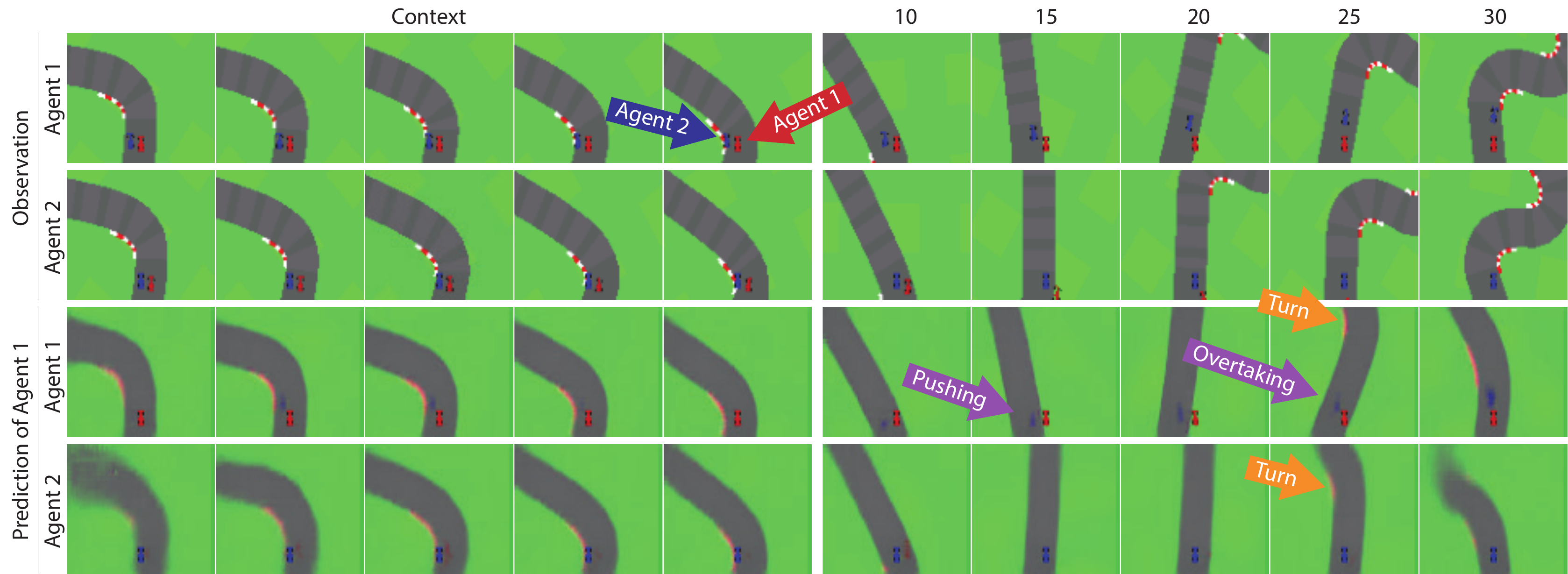}
\end{center}
\caption{\textbf{Open loop prediction:} Agent 1 (red) receives 5 observation frames as context and predicts the own view and the view of agent 2 (blue). We employ the learned dynamics to forward propagate 25 additional timesteps into the future without any further observations.}
\label{fig:observer_openloop}
\end{figure}

\paragraph{Imagining interaction sequences}
%
To facilitate efficient self-play in a learned world model, we require all agent states to be jointly propagated in a consistent manner. 
In Figure~\ref{fig:observer_openloop}, we provide 5 observations of agent 1 as context and investigate the model's ability to predict forward in time for 25 additional timesteps. Similarly to Figure~\ref{fig:observer}, we exclude ground-truth observations of agent 2. While actions of agent 1 are available for the full horizon, the learned policy predicts the actions of agent 2.
%

Referring to the context frames, we observe the benefit of recursively estimating states in agent 1's reconstruction of agent 2's viewpoint: while the reconstruction has high uncertainty in the first frame, accuracy improves with every new observation. Agent 1's reconstruction of their ego view is detailed from the start with the blue agent remaining blurry.
Moving beyond the context frames, the predicted views quickly diverge from the ground truth. This is expected as the agent is unable to see beyond the first turn. In the following imagination, the blue agent overtakes the red agent by forcing them off the track (Figure~\ref{fig:observer_openloop} prediction of agent 1, timestep 15). This is reasonable, as the blue agent leaves the left turn on the inside with an opportunity to push the red agent off the road. The imagined sequences feature detailed predictions of the track, including markings on an upcoming left turn (timestep 25).

Most importantly, the predictions of both agents remain consistent. The two agents' relative positions with each other and the track are in correspondence along the full horizon. Likewise, the predicted characteristics of a track including curbs and a left turn are consistent between both agents' predicted views (timestep 25).
Consistency is crucial as it allows to learn from imagined self-play in a world model. If the two predictions were to diverge, the impact of the ego actions on another agent's behavior could not be estimated. The outcome of the imagined games would be uncertain. The joint transition model helps keeping predictions consistent, as it allows for information exchange between both agents' latent states during forward propagation.

\section{Conclusion}
We present Deep Latent Competition (DLC), a novel reinforcement learning algorithm that learns competitive visual control policies through self-play in imagination.
%
The DLC agent can imagine interaction sequences in the compact latent space based on a multi-agent world model that combines a joint transition function with opponent viewpoint prediction.
%
%
The behavior is then optimized by back-propagating the analytic value gradients of these imagined game trajectories through the learned world model.
%
%
Experiments in a novel continuous visual control racing environment demonstrate that the DLC agent learns to make consistent multi-agent forward predictions. Optimizing competitive behaviors through imagined self-play based on these joint predictions yields an agent that performs superior to an agent that propagates ground truth observations separately.
%
%
In the future, we aim to deploy the DLC agent on hardware platforms to yield competitive racecar driving in the real world. Extending the framework to include more complex game-theoretic considerations in the forward predictions offers another intriguing avenue for future work.

\section*{Acknowledgments}
This work was supported in part by Qualcomm and Toyota Research Institute (TRI). This article solely reflects the opinions and
conclusions of its authors and not TRI, Toyota, or any other entity. We thank them for their support.
The authors further would like to acknowledge the MIT SuperCloud and Lincoln Laboratory Supercomputing Center for providing
HPC resources. We would also like to thank Danijar Hafner for open-sourcing the Dreamer agent.





{
\small 
\bibliography{references}
}
\newpage


\appendix

\section{Network Architectures}
\begin{table}[H]
\centering
 \caption{General network architectures of the underlying models. The transition model is joint in our implementation and the associated input and output variables correspond to joint representations (e.g. the previous action $a_{\tau-1}$ (6) can be interpreted as $a_{\tau-1}$ (2$
 \times$3)). We note that repeated layers have been condensed with Dense $\times$ $i$ referring to application of the same dense layer architecture $i$ times. The employed parameter abbreviations are referring to: a=activation, k=kernel, p=padding, s=stride.}
 \label{tbl:params}
  \begin{tabular}{l l l l}
   \toprule
    Layer Type $\qquad$ & Input (dimensions) & Output (dimensions) & Additional Parameters \\
   \toprule
    \mcTwo{Transition model (\textit{imagine 1-step})} & & \\
   \midrule
    Dense    & $s_{\tau-1,s}$ (60), $a_{\tau-1}$ (6) & $\text{fc}_{t,i}^1$ (600)   &  a=ELU        \\[1ex]
    GRU    & $\text{fc}_{t,i}^1$ (600), $s_{\tau-1,d}$ (400) & $\text{rs}_{\tau}$ (400),  $s_{\tau,d}$ (400)   &  a=tanh        \\[1ex]
    Dense    & $\text{rs}_{\tau}$ (400) & $\text{fc}_{t,i}^2$ (600)   &  a=ELU        \\[1ex]
    Dense    & $\text{fc}_{t,i}^2$ (600) & $\mu_{\tau,s}^{prior}$ (60), $\sigma_{\tau,s}^{prior}$ (60)   &  a=None        \\
   \toprule
    \mcTwo{Transition model (\textit{observe 1-step})} & & \\
   \midrule
    Dense    & $s_{\tau,d}$ (400), $z_{\tau}$ (2048) & $\text{fc}_{t,o}^1$ (600)   &  a=ELU        \\[1ex]
    Dense    & $\text{fc}_{t,o}^1$ (600) & $\mu_{\tau,s}^{post}$ (60), $\sigma_{\tau,s}^{post}$ (60)   &  a=None        \\
   \toprule
    \mcTwo{Encoder model} & & \\
   \midrule
    Conv2D    & obs (96, 96, 3) & cv1 (31, 31, 32)   &  a=ReLU, s=3, k=(4,4)        \\[1ex]
    Conv2D & cv1 (31, 31, 32) &  cv2 (14, 14, 64)   &  a=ReLU, s=2, k=(4,4)        \\[1ex]
    Conv2D      & cv2 (14, 14, 64) &  cv3 (6, 6, 128)   &  a=ReLU, s=2, k=(4,4)  \\[1ex]
    Conv2D      & cv3 (6, 6, 128) &  cv4 (2, 2, 256) & a=ReLU, s=2, k=(4,4) \\
   \toprule
    \mcTwo{Observation model} & & \\
   \midrule
    Dense     &  $s_{\tau,d}^i$ (200), $s_{\tau,s}^i$ (30) &  $\text{fc}_{o}^1$ (1, 1, 1024)   &  a=None      \\[1ex]
    Deconv2D    & $\text{fc}_{o}^1$ (1, 1, 1024) &  dc1 (5, 5, 128)   & a=ReLU, s=2, k=(5,5) \\[1ex]
    Deconv2D & dc1 (5, 5, 128) &  dc2 (13, 13, 64)   &  a=ReLU, s=2, k=(5,5)       \\[1ex]
    Deconv2D      & dc2 (13, 13, 64) &  dc3 (31, 31, 32) &  a=ReLU, s=2, k=(6,6), p=1       \\[1ex]
    Deconv2D      & dc3 (31, 31, 32) & dc4 (96, 96, 3)   &  a=ReLU, s=3, k=(6,6)      \\
   \toprule
    \mcTwo{Reward model} & & \\
   \midrule
    Dense     & $s_{\tau,d}^i$ (200), $s_{\tau,s}^i$ (30) &  $\text{fc}_{r}^1$ (400)   &  a=ELU      \\[1ex]
    Dense $\times$ 1     & $\text{fc}_{r}^{\{1\}}$ (400) &  $\text{fc}_{r}^{\{2\}}$ (400)   &  a=ELU      \\[1ex]
    Dense     & $\text{fc}_{r}^2$ (400) &  $\text{fc}_{r}^3$ (1)   &  a=ELU   \\
   \toprule
    \mcTwo{Value model} & & \\
   \midrule
    Dense     &  $s_{\tau,d}^i$ (200), $s_{\tau,s}^i$ (30) &  $\text{fc}_{v}^1$ (400)   &  a=ELU      \\[1ex]
    Dense $\times$ 2     & $\text{fc}_{v}^{\{1, 2\}}$ (400) &  $\text{fc}_{v}^{\{2, 3\}}$ (400)   &  a=ELU      \\[1ex]
    Dense     & $\text{fc}_{v}^3$ (400) &  $\text{fc}_{v}^4$ (1)   &  a=ELU   \\
   \toprule
    \mcTwo{Action model} & & \\
   \midrule
    Dense     &  $s_{\tau,d}^i$ (200), $s_{\tau,s}^i$ (30) &  $\text{fc}_{a}^1$ (400)   &  a=ELU      \\[1ex]
    Dense $\times$ 3     & $\text{fc}_{a}^{\{1, 2, 3\}}$ (400) &  $\text{fc}_{a}^{\{2, 3, 4\}}$ (400)   &  a=ELU      \\[1ex]
    Dense     & $\text{fc}_{a}^4$ (400) &  $\mu_{a}$ (3), $\sigma_{a}$ (3) &  a=ELU   \\
   \bottomrule
  \end{tabular}
\end{table}
Based on the general network architectures provided in Table~\ref{tbl:params}, we comment on how the two parts of the transition model integrate with each other and provide further details on each of the models.
\paragraph{Transition model}
The joint transition model follows the recurrent state space model (RSSM) architecture presented in~\cite{hafner2018learning, Hafner2020Dream}. The RSSM is extended to the multi-agent setting and propagates joint model states consisting of a deterministic and a stochastic component, respectively denoted by $s_{t,d}$ and $s_{t,s}$ at time $t$. The stochastic component $s_{t,s}$ is implemented via a diagonal Gaussian distribution and its derivation is provided in the next section. The transition model then predicts priors for the associated mean and standard deviation based on the previous model state and applied action (\textit{imagine 1-step}). In the presence of observations, prior estimates can be updated to posterior estimates (\textit{observe 1-step}). The transition model may then initialize its states by propagating posteriors based on a context sequence (\textit{imagine 1-step} and \textit{observe 1-step}) from which interactions can be imagined by propagating prior estimates (\textit{imagine 1-step}).

\paragraph{Encoder model} 
The encoder parameterization follows the architectural choices presented in~\cite{Ha2018}, where we adapt the convolutional layers to match the dimensionality of our observations. The agent leverages two encoders in parallel, one for generating latent vectors of the ego perspective and one for predicting latent vectors of the opponent perspective. Inputs are 96$\times$96 RGB image observations.
\paragraph{Observation model} 
The observation model follows the decoder architecture presented in~\cite{Ha2018}, where we adapt the transposed convolutional layers to match the dimensionality of our observations. The image observations of agent $i$ are reconstructed from the associated model states $s_{\tau, d}^i$ and $s_{\tau, s}^i$.
\paragraph{Reward and value model}
Rewards and values of agent $i$ are both predicted as scalar values from fully-connected networks that operate on the associated model states $s_{\tau, d}^i$ and $s_{\tau, s}^i$, similar to~\cite{Hafner2020Dream}.
\paragraph{Action model}
The action model follows~\cite{Hafner2020Dream}, where the predicted mean $\mu_{a}$ is rescaled and passed through a tanh to allow for saturated action distributions. It is combined with a softplus standard deviation based on $\sigma_{a}$ and the resulting Normal distribution is again squashed using a tanh~\cite{Hafner2020Dream, haarnoja2018soft}.

\paragraph{Order independence}
We note as an implementation detail that the transition distribution $q_\theta \left(s_{t}^1, s_{t}^2 | s_{t-1}^1, s_{t-1}^2, a_{t-1}^1, a_{t-1}^2\right)$ should be independent of the order in which agents are provided. Thus, reversing the ordering and considering $q_\theta \left(s_{t}^2, s_{t}^1 | s_{t-1}^2, s_{t-1}^1, a_{t-1}^2, a_{t-1}^1\right)$ should yield the same distribution. We achieve this by two passes through the learned transition model and subsequent averaging. In the second forward pass the agents' input order to the transition model is flipped.

\paragraph{Training parameters}
Most of the training parameters correspond to the implementation of \cite{Hafner2020Dream}, a state-of-the-art model-based RL algorithm for learning to plan in latent-space from image observations.
DLC trains every 1000 environment steps for 200 iterations with the Adam optimizer~\cite{kingma2014adam}. The batch size is set to 50.  The representation, value and actor model are respectively trained with learning rates 6e-4, 6e-4, and 8e-5. Gradients over the magnitude of 100 are clipped for all models. The prior $\sigma_{\tau,s}^{prior}$  and posterior $\sigma_{\tau,s}^{post}$  variance in the transition model are bounded from below to a minimum value of 0.1.
The model loss on true observations $J_{M,s_t^1, s_t^2}$ is weighted twice as much as the model losses on predicted opponent observations $J_{M,s_t^1, \tilde{s}_t^2}$ and $J_{M,\tilde{s}_t^1, s_t^2}$. Throughout, we use $\gamma=0.99$ and $\lambda=0.95$.
The model learning horizon is $L=50$ whereas the imagination horizon is $H=15$. Value and action models are trained on the same trajectory rollouts.

\paragraph{Environment details} 
To ensure generalization, we randomized both the color and initial position of all vehicles throughout training. We also evaluated (but did not use) penalization for driving in the backward direction which can occur after returning from a spin on the grass.


\section{Derivations}\label{app:model_deriv}
Our representation model is a latent variable model. That is, it can be written as
\begin{align*}
p_\theta(S_t | S_{t-1}, A_{t-1}, O_t)
  &= 
  p_\theta \left(
    s_{t}^1, s_{t}^2 | s_{t-1}^1, s_{t-1}^2, a_{t-1}^1, a_{t-1}^2, o_t^1, o_t^2
  \right)\\
  &= \int \int p_\theta \left(
    s_{t}^1, s_{t}^2 | s_{t-1}^1, s_{t-1}^2, a_{t-1}^1, a_{t-1}^2, z_t^1, z_t^2
    \right) \\
  &\qquad\qquad\qquad \cdot 
    p(z_t^1 | o_t^1) \ \cdot p(z_t^2 | o_t^2) 
    \ \mathrm{d} z_t^1\ \mathrm{d} z_t^2\ ,
\end{align*}
where $S_t:=\{s_t^1,\,s_t^2\}$ is the concatenation of both states and $A_{t-1}$, $O_{t-1}$ are defined analogously. Each agent $i$ maintains its own belief of the world, without having access to both observations. Thus, at test time it computes instead
\begin{align*}
p_\theta(S_t | S_{t-1}, A_{t-1}, o^i_t) 
    &= \int \int p_\theta \left(
        s_{t}^1, s_{t}^2 | s_{t-1}^1, s_{t-1}^2, a_{t-1}^1, a_{t-1}^2, z_t^i, \tilde{z}_t^{\neg i}
       \right)  \cdot q_\theta(z_t^i, \tilde{z}_t^{\neg i} | o_t^i) 
    \ \mathrm{d} z_t^i\ \mathrm{d} \tilde{z}_t^{\neg i}\ .
\end{align*}
Therefore, we never need to learn $p(z_t^i | o_t^i)$. Instead we directly learn the representation $q_\theta(z_t^i, \tilde{z}_t^{\neg i} | o_t^i) $. This is implemented as a convolutional encoder $f_E(o^i)$ making $q_\theta(z_t^i, \tilde{z}_t^{\neg i} | o_t^i) $ a Dirac distribution
$$
q_\theta(z_t^i, \tilde{z}_t^{\neg i} | o_t^i) 
= \delta \left(
  (z_t^i, \tilde{z}_t^{\neg i})-f_E(o_t^i)\right)\ .
$$

\newpage

\section{Additional Visualizations of Common Scenarios}
%
\begin{figure}[b!]
\begin{center}
    \includegraphics[width=0.62\linewidth]{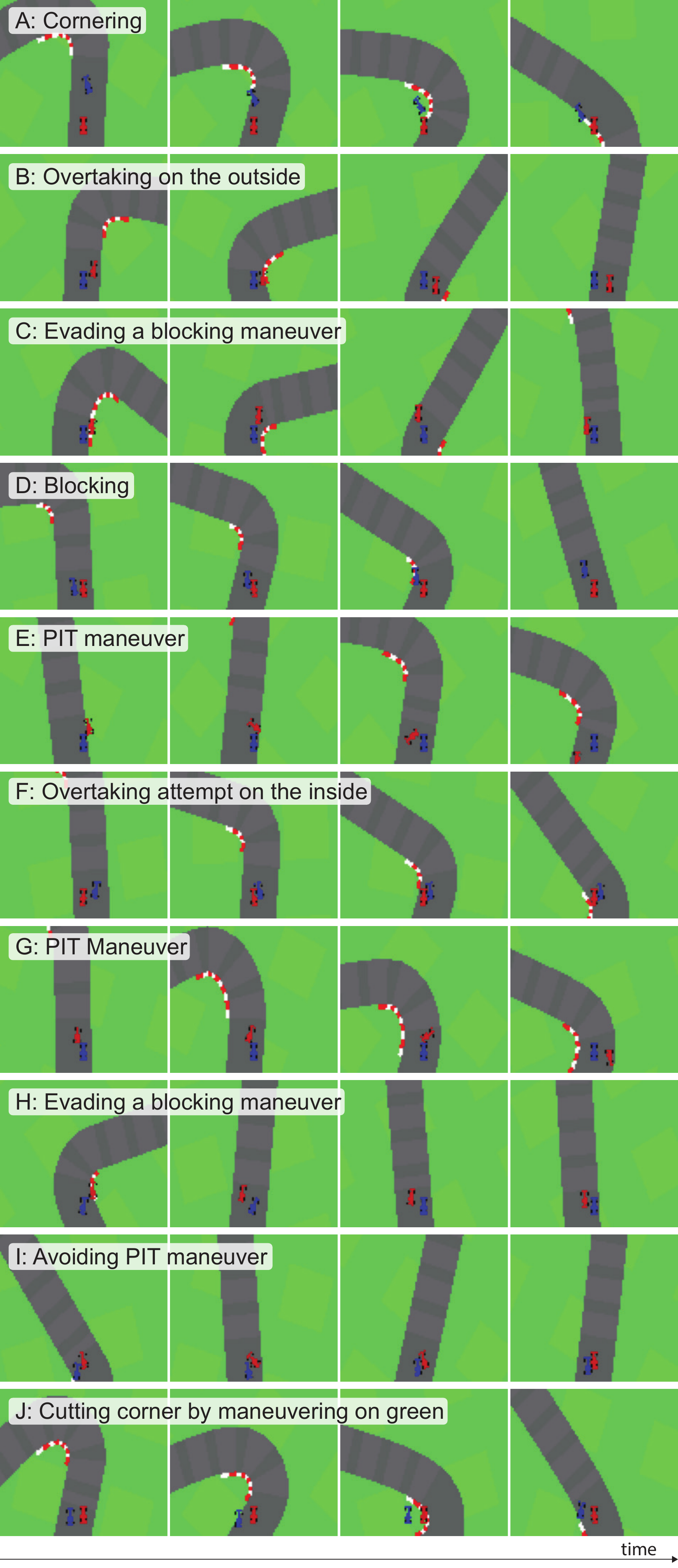}
\end{center}
\caption{The agent has learned to leverage a large variety of racing skills towards their competitive advantage. The skills include cornering, blocking, overtaking, and forcing others off the road.}
\label{fig:scenarios_appendix2}
\end{figure}
\begin{figure}
\begin{center}
    \includegraphics[width=0.62\linewidth]{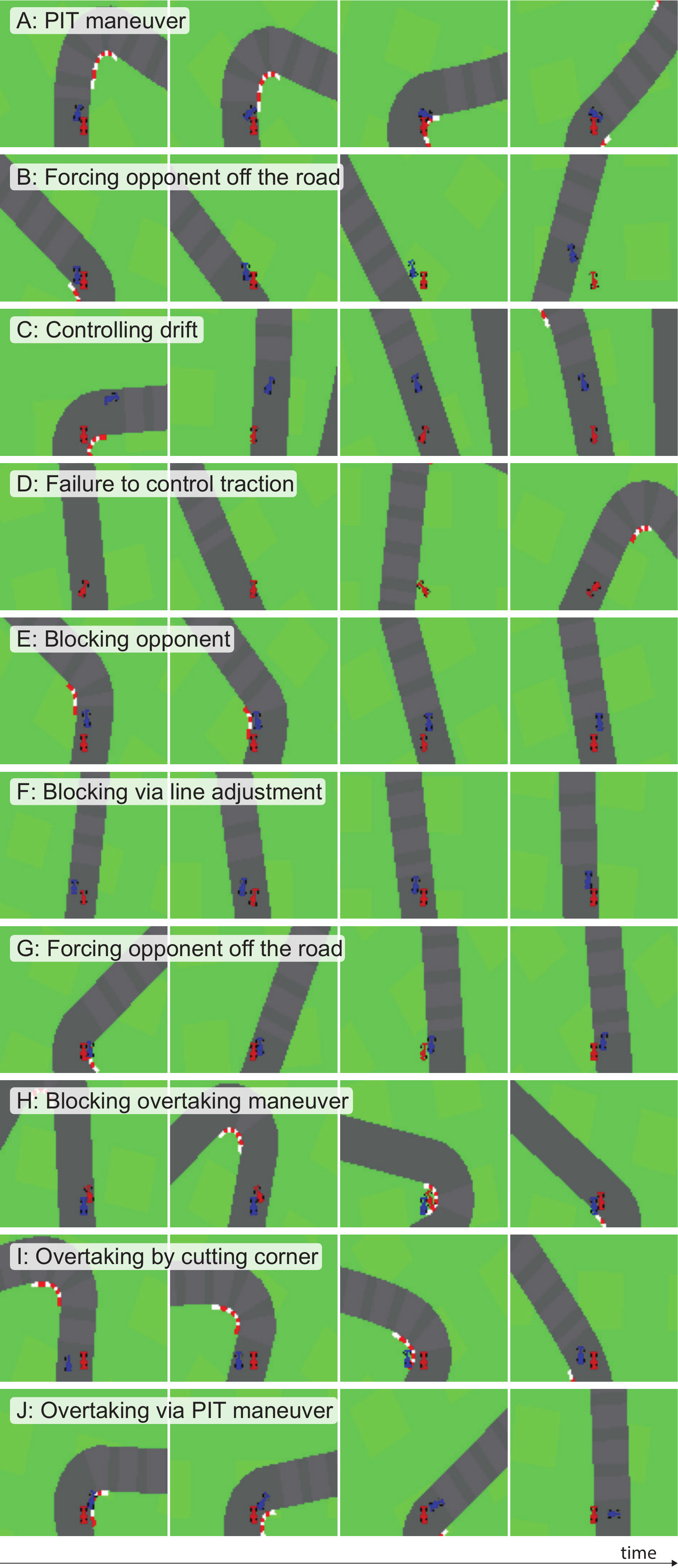}
\end{center}
\caption{During a race, agents have to drive at the dynamical limits of handling to move as fast as possible along the track. While they don't receive penalties for leaving the road, moving onto the slippery grass increases the risk of spinning out.}
\label{fig:scenarios_appendix1}
\end{figure}

\end{document}